\documentclass[10pt, a4paper]{article}

\usepackage[final]{lrec2026} 
\usepackage{booktabs}
\usepackage{multirow} 
\usepackage{amsmath} 
\usepackage{amssymb}
\usepackage{xcolor,colortbl}
\usepackage{booktabs}

\title{HatePrototypes: Interpretable and Transferable Representations for Implicit and Explicit Hate Speech Detection}

\name{Irina Proskurina\textsuperscript{1,2}  \quad Marc-Antoine Carpentier\textsuperscript{3}    \quad Julien Velcin \textsuperscript{3,4}  \\}

\address{\textsuperscript{1}Laboratoire Hubert Curien, UMR CNRS 5516, Saint-Etienne\\
\textsuperscript{2}Université Lumière Lyon 2, Université Claude Bernard Lyon 1, ERIC, Lyon\\ 
\textsuperscript{3}École Centrale de Lyon\\
\textsuperscript{4}LIRIS CNRS UMR 5205\\}

\abstract{
Optimization of offensive content moderation models for different types of hateful messages is typically achieved through continued pre-training or fine-tuning on new hate speech benchmarks. However, existing benchmarks mainly address explicit hate toward protected groups and often overlook implicit or indirect hate, such as demeaning comparisons, calls for exclusion or violence, and subtle discriminatory language that still causes harm. While explicit hate can often be captured through surface features, implicit hate requires deeper, full-model semantic processing. In this work, we question the need for repeated fine-tuning and analyze the role of \textbf{HatePrototypes}, class-level vector representations derived from language models optimized for hate speech detection and safety moderation. We find that these prototypes, built from as few as 50 examples per class, enable cross-task transfer between explicit and implicit hate, with interchangeable prototypes across benchmarks. Moreover, we show that parameter-free early exiting with prototypes is effective for both hate types. 
We release the code, prototype resources, and evaluation scripts to support future research on efficient and transferable hate speech detection.
 \\ \newline \Keywords{hate speech, efficiency, transfer, prototypes, early exiting} }

\begin{document}

\maketitleabstract

\section{Introduction}\label{sec:introduction}
The impact of online hate comments and their harmful consequences spans a wide range of effects, from individual hate crimes and psychological trauma to the disruption of group discussions, distortion of community norms, distraction from the main post content, and discouragement of user participation \cite{muller2021fanning,lees2022new}, with teenagers being particularly at risk\footnote{\href{https://www.microsoft.com/en-us/digitalsafety/research/global-online-safety-survey}{microsoft.com/en-us/digitalsafety}}.
Existing hate detection benchmarks include messages and moderated content describing attacks on individuals or groups based on protected characteristics such as race, religion, or gender \citep{pachinger2023toward,albanyan2023finding}.
Language models (LMs) are widely used for online platform moderation and for providing rewriting suggestions for potentially hateful content \cite{waseem-hovy-2016-hateful,dale-etal-2021-text}.

Although LMs fine-tuned to classify text as hateful perform well on in-domain hate messages, they exhibit two main limitations in (1) real-world social media moderation \cite{tonneau-etal-2025-hateday} and (2) real-time settings \citep{oikawa2022stacking}.

Research on the real-world application of LM-based hate speech detectors shows that current systems focus primarily on explicit hate, often relying on slur-based features, either rule-based or model-driven \cite{schmidt2017survey}. This causes specific limitations in (i) out-of-domain multilingual settings \cite{tonneau-etal-2025-hateday}, (ii) implicit hate detection without explicit lexical cues \cite{sap-etal-2020-social,elsherief2021latent,sridhar2022explaining}, and (iii) misclassification of neutral examples as hateful \cite{diaz2021double,hartvigsen-etal-2022-toxigen}.
\citet{wiegand2019detection} highlight the problem of poor out-of-domain performance, that is, low transferability caused by training data biases in hate and abusive text detection.
To improve out-of-domain performance, existing methods propose training data augmentation \cite{kim-etal-2022-generalizable,jin-etal-2024-gpt} and contrastive learning approaches that leverage shared hate implications or group representations across datasets \cite{kim-etal-2022-generalizable,ocampo2020detecting,ahn-etal-2024-sharedcon}.

Research on real-time hate detection focuses on the out-of-domain live performance of models on streaming platforms and user platform interactional dialogues \cite{inan2023llama,yang2023toxbuster}.
To reduce the latency of classification models, various acceleration techniques have been proposed \cite{treviso2023efficient}. Recent works discuss early exiting techniques designed to classify shorter and simpler instances faster by performing classification at earlier LM layers \citep{rahmath2024early}. Early exiting has also been applied to accelerate other classification tasks such as sentiment analysis \citep{xin-etal-2020-deebert,liu2021towards,elhoushi2024layerskip}.
Motivated by the limitations highlighted in both research directions, this work studies out-of-domain transfer in LMs without the need for fine-tuning, using \textbf{HatePrototypes}, vector representations of hate classes, evaluated across cross-domain benchmarks.
While prior studies have examined cross-task and cross-domain transfer and questioned the extent to which language models encode shared representations of hate-related semantics, to the best of our knowledge, no prior work (1) employs prototype-based classification to analyze the transferability of LMs fine-tuned on implicit versus explicit hate benchmarks, or (2) explores how such transfer can be achieved without fine-tuning between implicit and explicit hate speech tasks.
We further analyze prototypes constructed layer-wise to guide early exiting in models for implicit and explicit hate detection.

Overall, our contributions are as follows:
(1) We analyze the role of HatePrototypes in the transferability of LMs optimized for implicit hate detection, explicit hate detection, or general safety moderation, and find significant performance improvements in prototype-based transfer between models.
(2) We show that HatePrototypes are transferable between implicit and explicit hate messages, with consistent findings across two model families.
(3) We further explore early exiting in models fine-tuned on implicit and explicit hate speech tasks, demonstrating how layer wise prototype construction can enhance efficiency and performance.

\noindent{\textcolor{red}{\textbf{Content warning:} This article contains illustrative examples of hateful content.}}

\section{Related Work}\label{sec:related-work}

\paragraph{Transferability in Hate Speech Detection}
Despite strong in-domain performance, language models for hate speech detection often fail to transfer across datasets, platforms, or categories of abuse \citep{pachinger2023toward, khurana-etal-2022-hate}.
Early studies demonstrate that differences in dataset design and annotation practices outweigh architectural choices in determining generalization performance \citep{arango2019hate, vidgen-derczynski-2020}.
Subsequent cross-corpus and functional evaluations confirm these limitations, revealing persistent errors on simple linguistic variations such as negation, euphemisms, and counter-speech \citep{rottger-etal-2021-hatecheck, tonneau-etal-2025-hateday}.
Later work aims to improve transferability through dataset-centered techniques, including data augmentation, multi-dataset training, and granular multi-label or facet-based hate type classification \citep{mathew2021hatexplain, inan2023llama, leite-etal-2023-noisy}, as well as training-centered enhancements such as contrastive learning to align latent representations \citep{macavaney2019hate, kim-etal-2022-generalizable, jafari2023fine}.
However, performance gains remain inconsistent, particularly when transferring between explicit and implicit hate domains \citep{ocampo2020detecting}.
At the same time, findings from research on implicit social bias indicate that pre-trained LMs systematically assign higher likelihood to stereotypical associations than to anti-stereotypical alternatives, demonstrating the encoding of implicit biases in the model representations \citep{nangia-etal-2020-crows,nadeem-etal-2021-stereoset,zhao-etal-2025-explicit}.
Recently, \citet{ahn-etal-2024-sharedcon} introduced \textsc{SharedCon}, a clustering-based contrastive learning framework for implicit hate speech classification that leverages shared semantic structures to enhance generalization.

\paragraph{Early Exiting}
Early exiting, or anytime prediction, is an acceleration technique in which an input sample $x$ can be classified at intermediate layers of a language model, producing a prediction $y(x)$ without processing the entire model.
Early works on exiting propose adding additional classification heads at intermediate layers, which are used at inference time in a gated manner based on prediction confidence (DeeBERT; \citealt{xin-etal-2020-deebert}), and with confidence stabilization constraints (PABEE; \citealt{zhou2020bert}).
Later works adopt a teacher-student framework for gated exits (FastBERT; \citealt{liu2020fastbert}) and extend multi-layer exiting during pre-training (ElasticBERT; \citealt{liu2021towards}).
Originally proposed for classification tasks, ongoing work adapts early exiting to generative applications at the token level \citep{tang-etal-2024-deed, bae-etal-2023-fast}.
Recent studies explore representation-driven or \textit{prototype-based} distance-aware early exiting for language and vision models \citep{he20243, gormez20222, snell2017prototypical, xie-etal-2023-proto}.
To the best of our knowledge, no prior work has focused on prototype-based early exiting for hate speech detection or on studying its effect on transferability.

\section{Methodology}\label{sec:methodology}

We use training subsets of multiple hate speech benchmarks to construct \textit{HatePrototypes}, class centroids representing the mean embedding of each of the hate and non-hate classes.
These prototypes are further used for cross-task transfer analysis and layer-wise classification for model early exiting.

Let $\mathcal{D}=\{(x_i,y_i)\}_{i=1}^N$ be the training corpus with labels non-hate and hate $y_i\in\mathcal{C}=\{0,1\}$.
For each class $c\in\mathcal{C}$, let $\mathcal{D}_c$ denote the class-specific subset.

Let the model consist of $L$ transformer layers with hidden size $d$. 
For an input sequence $x$, we denote by $h^{(\ell)}(x) \in \mathbb{R}^{d}$ 
the sequence-level representation extracted at layer $\ell$. 
For encoder models such as BERT, we use the hidden state of the \texttt{[CLS]} token at layer $\ell$.
For decoder-based models such as OPT and LLaMA, we use the hidden state of the last non-padding token.
For each class $c\in\{0,1\}$ and layer $\ell$, we construct a class prototype by averaging the training representations of that class:
\begin{equation}\label{eq:hate-prototypes}
\mu_c^{(\ell)} \;=\; \frac{1}{|\mathcal{D}_c|}\sum_{(x,y)\in\mathcal{D}_c} h^{(\ell)}(x),
\qquad
\mu_c^{(\ell)}\in\mathbb{R}^d.
\end{equation}

At inference time, for a new input $x$, we measure its similarity $s^{(\ell)}(x)$ to both class prototypes at layer $\ell$ as:
\begin{equation}\label{eq:similarity}
s_c^{(\ell)}(x)=\tilde h^{(\ell)}(x)^\top \tilde \mu_c^{(\ell)},
\end{equation}
where both representations are $\ell_2$ normalized:
\[
\tilde h^{(\ell)}(x) \;=\; \frac{h^{(\ell)}(x)}{\|h^{(\ell)}(x)\|_2},
\qquad
\tilde \mu_c^{(\ell)} \;=\; \frac{\mu_c^{(\ell)}}{\|\mu_c^{(\ell)}\|_2}.
\]
Let $s^{(\ell)}_{(1)}\ge s^{(\ell)}_{(2)}$ be the largest and second-largest values among $\{s_c^{(\ell)}(x)\,:\,c\in\{0,1\}\}$. 
To enable early exiting during inference, we compute the \emph{margin} (the per-sample confidence gap) at each layer and stop the forward pass at the first layer where the difference between the largest and second-largest similarity scores satisfies:
\begin{equation}\label{eq:main-exit}
m^{(\hat{\ell})}(x) \;\ge\; \delta,
\end{equation}
where $\delta>0$ is a fixed margin threshold controlling the efficiency-accuracy trade-off. 
If no layer meets this condition, inference proceeds through all $L$ layers and the prediction from the final layer $L$ is used as a model's output.

\section{Experimental Setup}\label{sec:exp-setup}

\paragraph{Models}
To compare how different architectures encode hate speech, we use two models of comparable size: the encoder BERT-base\footnote{\href{https://huggingface.co/google-bert/bert-base-cased}{hf.co/bert-base-cased}}~\cite{devlin2019bert} with 109M parameters and the decoder OPT-125M\footnote{\href{https://huggingface.co/facebook/opt-125m}{hf.co/facebook/opt-125m}}~\citep{zhang2022opt} with 125M parameters.  
OPT is pre-trained for causal language modeling with a 50k-token vocabulary, while BERT is pre-trained for masked language modeling with a 29k-token vocabulary.  
Despite architectural differences, both are case-sensitive, have 12 layers and 12 attention heads, and share a hidden size of 768, making them directly comparable.  
For experiments involving guardrail models designed for safety moderation in model generations, we use Llama-Guard-3-1B\footnote{\href{https://huggingface.co/meta-llama/Llama-Guard-3-1B}{hf.co/meta-llama/Llama-Guard-3-1B}} and BLOOMZ-Guardrail-3B\footnote{\href{https://huggingface.co/cmarkea/bloomz-3b-guardrail}{hf.co/cmarkea/bloomz-3b-guardrail}}.

\paragraph{Benchmarks}
For implicit hate detection, we use the Implicit Hate Corpus (IHC; \citetlanguageresource{elsherief2021latent}), which contains social media posts expressing prejudice or hostility toward protected groups through indirect or coded language, such as sarcasm, euphemism, or insinuation, and the Social Bias Inference Corpus (SBIC; \citetlanguageresource{sap-etal-2020-social}), which captures non-explicit statements that imply social stereotypes or biased assumptions about demographic groups. 
For explicit hate detection, we use the Offensive Language Identification Dataset (OLID; \citetlanguageresource{zampieri2019predicting}), designed for classifying group-targeted offense and insults, and HateXplain (\citetlanguageresource{mathew2021hatexplain}), which contains explicitly hateful and offensive posts annotated with targeted groups and rationales. 
All datasets are primarily sourced from Twitter, with HateXplain and SBIC additionally incorporating posts from Gab. 
Dataset statistics are presented in \autoref{tab:hate_speech_benchmarks}.

\begin{table}[t]
\centering
\footnotesize
\resizebox{0.45\textwidth}{!}{%
\begin{tabular}{cccc}
\toprule
 \textbf{Benchmark} & \textbf{$\bar{T}$} & \textbf{Hate \%} & \textbf{N Sent. (Train./Test)} \\
 \midrule
SBIC  & 28.10 & 68.32\% & 35,424/4,691 \\
IHC  & 20.41 & 34.82\% & 14,273/3,059 \\
\midrule
OLID & 32.10 & 31.21\% & 9,268/860  \\
HateXplain & 27.81 & 59.36\% & 15,379/1,924  \\
\bottomrule
\end{tabular}}
\caption{Overview of benchmarks for hate speech classification. $\bar{T}$ denotes the average length of tokenized texts; Hate~\% indicates the proportion of hate-class examples. 
SBIC and IHC are implicit hate benchmarks, while OLID and HateXplain are explicit hate.}
\label{tab:hate_speech_benchmarks}
\end{table}

We fine-tune BERT and OPT models on the training splits of the benchmarks for 3 epochs, with a learning rate of $1\times10^{-5}$ and a batch size of 64.
To address class imbalance, we use weighted cross-entropy loss with class-specific ratios computed from the training data of each benchmark.
Unless otherwise specified, all experiments are run with 10 random seeds on a single NVIDIA A100 GPU (80~GB).

\section{Prototypes for Task Transfer}\label{sec:results-prototypes-transfer}

In this section, we present the results of using prototypes to transfer knowledge between different tasks. Our experiments involve three datasets: one for fine-tuning the model, one for creating the prototypes, and one for testing performance. We investigate two types of transfer.

In the first case, referred to as \textit{cross-domain transfer}, the prototypes and evaluation samples come from the same dataset, while the evaluated models have been fine-tuned on different datasets. This setup examines how well prototypes from a given domain can be used across models trained on other domains.

In the second case, referred to as \textit{prototype-based transfer}, the model is fine-tuned and evaluated on the same dataset, but the prototypes are built from a different dataset. This setup tests how well the learned representations generalize across domains when the prototype source differs from the target data.
In all experiments, prototypes are extracted from the final encoder layer of the model, and each sample is assigned the class label corresponding to the highest similarity score, as defined in Eq.~\eqref{eq:similarity}.
We use the training subsets to construct the prototypes and the test data for evaluation.
\begin{table}[!t]
\centering
\scriptsize
\resizebox{0.48\textwidth}{!}{%
\begin{tabular}{lcccc} 
\toprule
\textbf{FT-Eval.}  &
\textbf{Acc.} & \textbf{$\Delta$ Acc.} & \textbf{F1} & \textbf{$\Delta$ F1} \\
\midrule
\multicolumn{5}{c}{\textbf{BERT-base}} \\
\midrule
HX-HX & 77.39$^{\phantom{*}}$ & -1.10 & 77.12 & -0.83 \\
HX-OLID & 75.58$^{*}$ & +2.62 & 67.39 & +20.42 \\
HX-IHC & 63.57$^{*}$ & -4.37 & 61.48 & +3.32 \\
HX-SBIC & 76.61$^{*}$ & +32.38 & 71.38 & +28.02 \\
\midrule
OLID-HX & 61.84$^{*}$ & -2.11 & 61.37 & +3.39 \\
OLID-OLID & 83.51$^{\phantom{*}}$ & +0.45 & 79.91 & +0.16 \\
OLID-IHC & 59.06$^{*}$ & +0.39 & 57.18 & +0.85 \\
OLID-SBIC & 67.16$^{\phantom{*}}$ & -0.60 & 65.15 & -0.47 \\
\midrule
IHC-HX & 62.98$^{*}$ & +1.12 & 62.51 & +3.71 \\
IHC-OLID & 68.10$^{*}$ & +15.39 & 60.77 & +9.75 \\
IHC-IHC & 75.45$^{\phantom{*}}$ & +0.60 & 73.78 & +0.26 \\
IHC-SBIC & 75.98$^{\phantom{*}}$ & +2.08 & 68.61 & +4.04 \\
\midrule
SBIC-HX & 68.02$^{*}$ & +5.34 & 67.31 & +16.75 \\
SBIC-OLID & 78.60$^{*}$ & +6.01 & 72.78 & +3.32 \\
SBIC-IHC & 61.61$^{*}$ & +7.38 & 60.66 & +6.53 \\
SBIC-SBIC & 85.63$^{*}$ & +0.90 & 82.37 & +0.20 \\
\midrule
\multicolumn{5}{c}{\textbf{OPT-125M}} \\
\midrule
HX-HX & 78.27$^{\phantom{*}}$ & -0.73 & 77.86 & -0.60 \\
HX-OLID & 75.62$^{*}$ & +1.78 & 67.51 & +15.48 \\
HX-IHC & 68.62$^{*}$ & -0.29 & 66.89 & +6.34 \\
HX-SBIC & 77.49$^{*}$ & +30.04 & 66.78 & +19.87 \\
\midrule
OLID-HX & 63.50$^{*}$ & -0.99 & 63.00 & +3.99 \\
OLID-OLID & 84.68$^{\phantom{*}}$ & +0.52 & 81.19 & +0.65 \\
OLID-IHC & 63.35$^{\phantom{*}}$ & +3.81 & 61.91 & +4.23 \\
OLID-SBIC & 73.39$^{*}$ & +7.90 & 70.05 & +6.11 \\
\midrule
IHC-HX & 67.46$^{*}$ & +5.72 & 67.04 & +8.22 \\
IHC-OLID & 70.62$^{*}$ & +23.89 & 64.61 & +18.16 \\
IHC-IHC & 78.02$^{*}$ & +0.44 & 75.91 & -0.00 \\
IHC-SBIC & 76.62$^{\phantom{*}}$ & +3.64 & 65.15 & +3.78 \\
\midrule
SBIC-HX & 69.33$^{*}$ & +5.31 & 68.71 & +16.76 \\
SBIC-OLID & 77.91$^{*}$ & +7.83 & 74.47 & +6.09 \\
SBIC-IHC & 67.87$^{*}$ & +15.78 & 66.91 & +15.38 \\
SBIC-SBIC & 85.82$^{*}$ & -0.38 & 81.22 & -2.44 \\
\bottomrule
\end{tabular}}
\caption{Accuracy and macro-F1 (\%) on evaluation benchmarks for cross-domain transfer with prototypes. 
Prototypes for classification are derived from fine-tuned (FT) models using the training data of the corresponding Eval dataset. 
$\Delta$ indicates the difference relative to the fine-tuned Eval baseline without prototypes. 
Statistically significant differences ($p<0.01$, paired $t$-test) are marked with $^*$.}
\label{tab:bert_opt_transfer}
\end{table}

\subsection{Cross-domain prototype transfer}\label{sec:encoder-transfer-prototypes}

We evaluate cross-domain classification performance with prototypes compared to fine-tuned baselines. 
Prototype centroids are computed from the last-layer hidden states of the models using the training set of the target classification benchmark Eq.\eqref{eq:hate-prototypes} with 500 examples per class.
To evaluate the effect of prototype size, we additionally examine smaller subsets (5–200 examples per class) in \S\ref{sec:prototype-selection}. 
We report fine-tuning domain (FT) - evaluation domain (Eval.) results in \autoref{tab:bert_opt_transfer}. 
Overall, prototypes significantly boost the performance of BERT- and OPT-based fine-tuned models across all cross-domain settings. 
The largest gains, relative to the FT baseline, for BERT occur when transferring from a HateXplain-tuned model to OLID (+20.42 F1) and to SBIC (+28.02 F1). 
For OPT, the biggest improvements are observed when transferring from a HateXplain-tuned model to SBIC (+19.87 F1) and from an IHC-tuned model to OLID (+18.16 F1) relative to the classification-head baseline. 
In all four cases, prototype-based performance approaches that of the classification-head baseline on FT data, with no statistically significant differences across seeds, except for OPT fine-tuned on both implicit benchmarks and BERT fine-tuned on SBIC (SBIC-SBIC and IHC-IHC rows marked with $^*$ in \autoref{tab:bert_opt_transfer}).

Regarding in-domain performance versus transfer performance, fine-tuned BERT and OPT achieve the highest scores on OLID and SBIC (F1$\approx$80-82); however, prototypes from these models do not yield the largest cross-domain F1-score gains. 
We further discuss whether this performance can be improved with prototype selection.

\paragraph{Training size impact}
We study the impact of fine-tuning data size and class imbalance on prototype performance by fine-tuning models under two conditions:
(1) equal-size datasets, subsampled to $2 \times \min(|\mathcal{D}_c|)$ examples (4,000 for OLID), and
(2) proportionally stratified datasets with 8,000 examples, preserving the original class imbalance.
We test for significant differences in model predictions relative to prototype-based classification on the fine-tuned test data for each seed using a two-sided paired $t$-test ($\alpha=0.01$).

Under balanced sampling, no statistically significant effects are observed across seeds. This result is expected, as we use weighted cross-entropy with class-balance ratios when fine-tuning on the full training sets. We do not report these results, as they are equivalent to those obtained from models trained on the full data.
Under reduced training size, stratified sampling yields significant differences compared to the prototype-based in-domain performance in two FT-Eval. pairs, when using prototypes from models fine-tuned on HateXplain and OLID to classify SBIC. For BERT fine-tuned on HateXplain and evaluated on SBIC, accuracy decreases to 74.00 (-2.60) and macro-F1 to 67.91 (-3.47) compared to the prototype-based in-domain performance. For OPT fine-tuned on OLID and evaluated on SBIC, accuracy increases to 74.95 (+1.56) and macro-F1 to 70.89 (+0.84) relative to the same baseline.

\begin{figure}[t]
\centering
  \includegraphics[width=0.9\columnwidth]{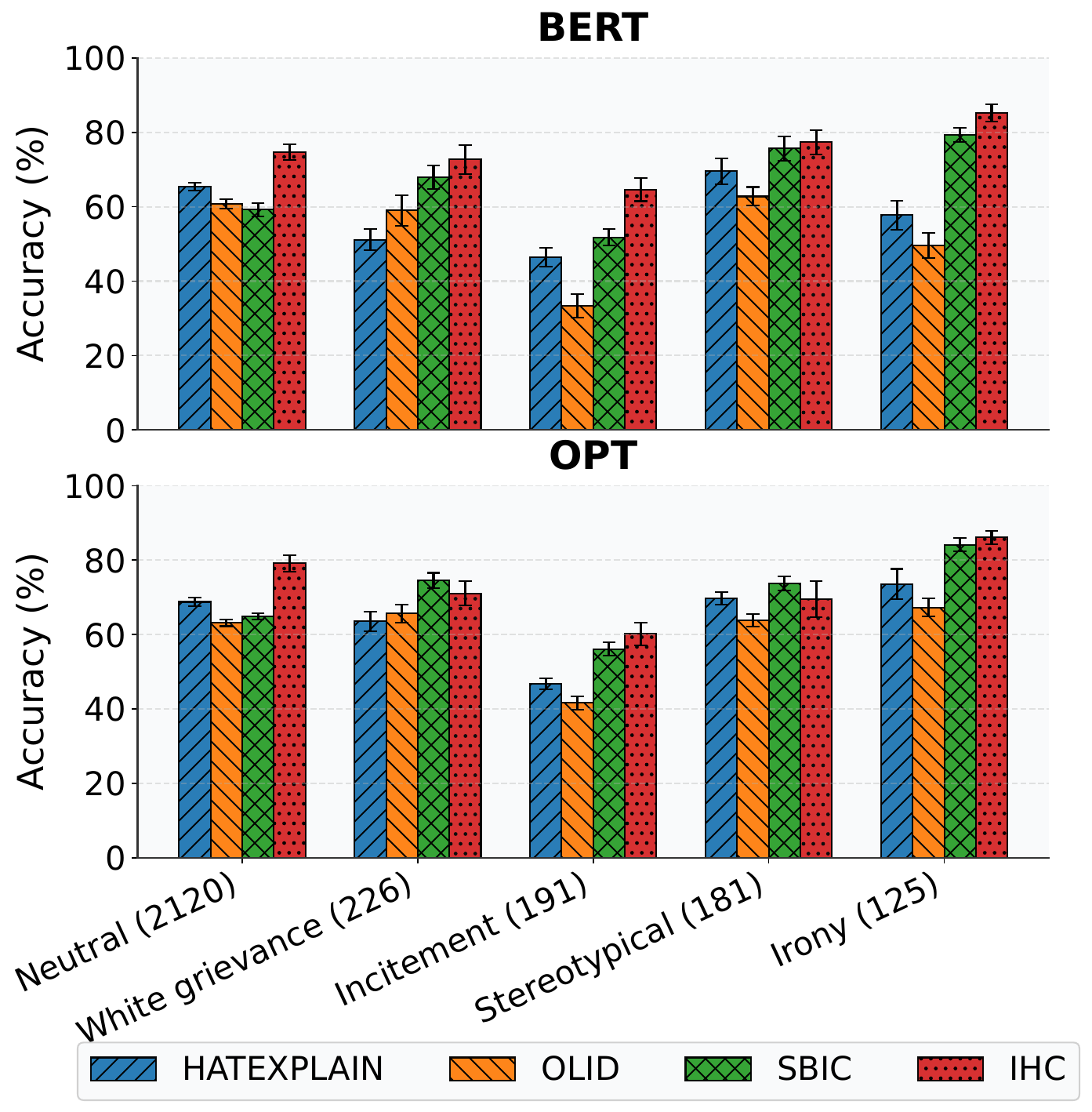}
\caption{Accuracy on IHC implicit hate types and neutral class obtained with IHC prototypes from models fine-tuned on 4 considered benchmarks.}
  \label{fig:bert-opt-grained}
\end{figure}

\paragraph{Fine-grained performance analysis}
To better understand the transfer performance, we analyze results across different hate types using fine-grained annotations from the IHC dataset. 
We focus on four frequent categories of implicit hate and the neutral class in IHC: \emph{white grievance}, \emph{incitement}, \emph{group stereotypes}, and \emph{irony}. 
The evaluation setting follows the previous setup: prototypes are extracted from the IHC training set using models fine-tuned on HX, OLID, SBIC, or IHC, and then used to classify IHC instances. 
We report accuracy scores averaged across runs in \autoref{fig:bert-opt-grained}. 
Overall, OPT and BERT exhibit similar performance across categories, with SBIC-tuned models performing on par with IHC-tuned models. 
The lowest accuracy is observed for the \emph{incitement} category, where accuracy for models fine-tuned on datasets other than IHC ranges between 40\% and 58\%.

To better understand these results, we conduct a qualitative analysis of examples that are consistently misclassified by BERT models fine-tuned on HateXplain and OLID across all seeds.  
For the irony category, misclassifications most often occur in examples featuring question-answer constructions or ironic framing, where the derogatory meaning is implied rather than expressed through overtly negative lexical cues.
Representative cases include \textit{“How do you call a Black man? You call his cell number.”}, \textit{“What do you get when you add 1 + 1? The average Ethiopian lifespan.”}. 
For incitement category, we observe two main patterns. The first pattern involves imperative or exhortative constructions, such as \textit{“Time to stop white flight \& start the white fight!”}. The second pattern encodes hostility or a call to action framed as neutral or even positive. These indirect expressions often rely on presupposed threats or appeals to group solidarity, as in \textit{“White Alabama is counting on you.”}, \textit{“Finns protest against increasing numbers of non-white invaders.”}, and \textit{“If other races can have self-care, whites can too.”}

\begin{figure}[t]
\centering
  \includegraphics[width=0.95\columnwidth]{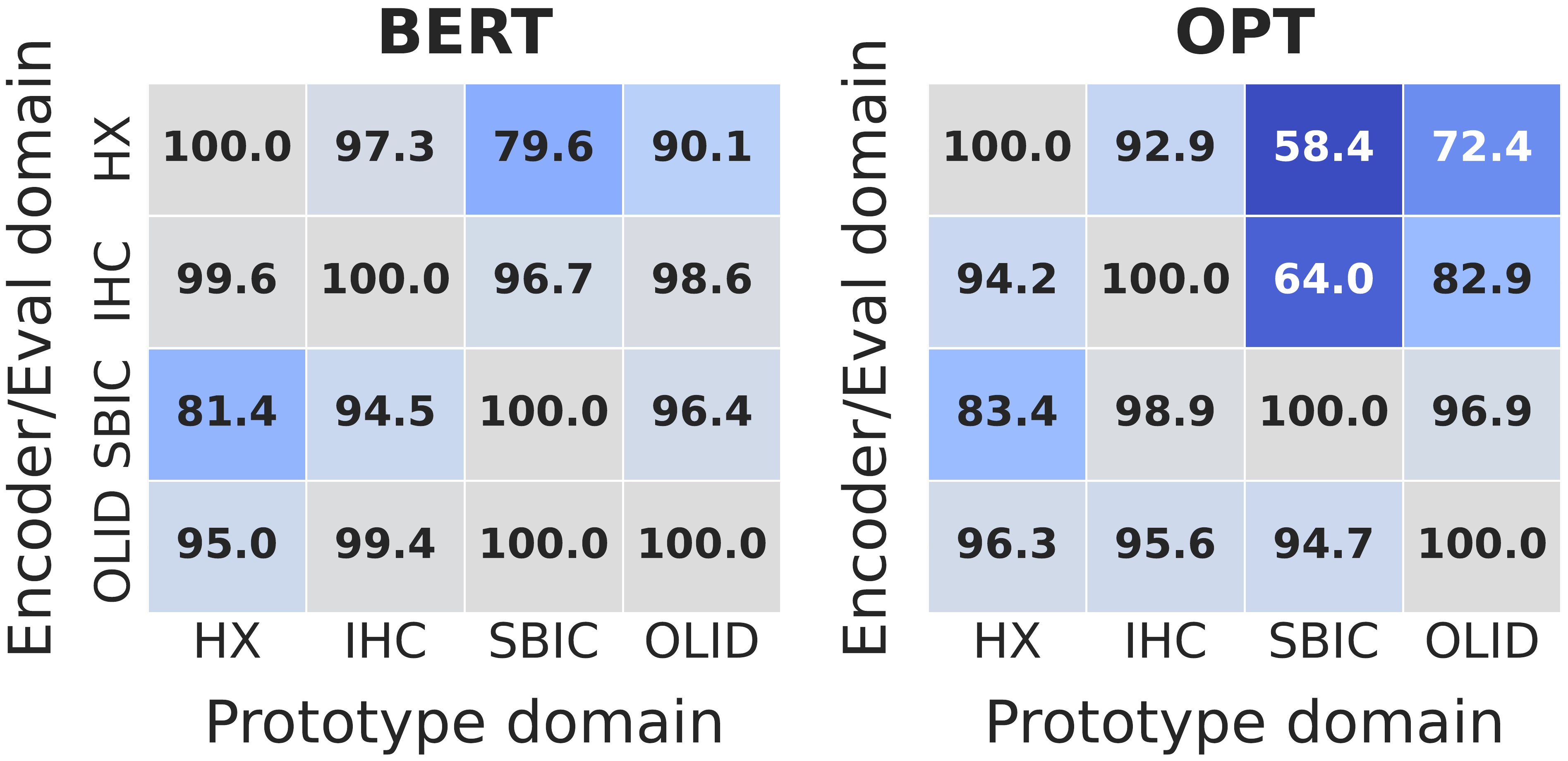}
\caption{Prototype selection results: relative cross-dataset transfer F1-scores for BERT and OPT models. 
Values show how much of each model’s in-domain performance carries over to other datasets.}
  \label{fig:prototype-transfer}
\end{figure}

\subsection{Prototype Transfer}\label{sec:prototypes-transfer}

Since in-domain training data may be unavailable in practice, we further test whether prototypes constructed from other datasets can be used to classify the target domain.  
In this setting, we evaluate classification performance on the same benchmarks used to fine-tune the encoders, while constructing prototypes from the training data of \emph{other benchmarks}. To quantify the proportion of in-domain performance that transfers, we report the relative macro-F1 with respect to the fine-tuned (FT) performance on the same data:$
\frac{F1\!\left(X \mid \mathrm{proto}(Y)\right)}{F1\!\left(X \mid \mathrm{proto}(X)\right)},$
where \(X\) denotes the encoder/evaluation domain and \(Y\) the prototype domain. 
Results across encoders and prototype domains are presented in \autoref{fig:prototype-transfer}.

Overall, prototypes constructed from other datasets achieve performance close to the fine-tuned (FT) baselines on their respective evaluation domains. The highest relative transfer is observed for the OLID-tuned model, which retains 95-100\% of its in-domain FT macro-F1 when classifying data using prototypes derived from other benchmarks. The lowest relative performance is obtained when prototypes constructed from the implicit SBIC dataset are applied to the explicit HateXplain domain.  

We also observe notable differences between OPT and BERT models. 
BERT maintains substantially higher relative macro-F1 scores compared to OPT on the implicit pair IHC-SBIC (96.7 vs.\ 64.0) and on the explicit pair HX-OLID (90.1 vs.\ 72.4).

On average, prototypes derived from the implicit IHC training set yield the highest relative macro-F1 across evaluation domains, while prototypes from the explicit HateXplain dataset consistently rank second across encoders.  
These results indicate that, despite domain differences, implicit benchmarks such as IHC can be used to construct prototypes for the classification of explicit domains.

\begin{figure}[t]
\centering
  \includegraphics[width=0.9\columnwidth]{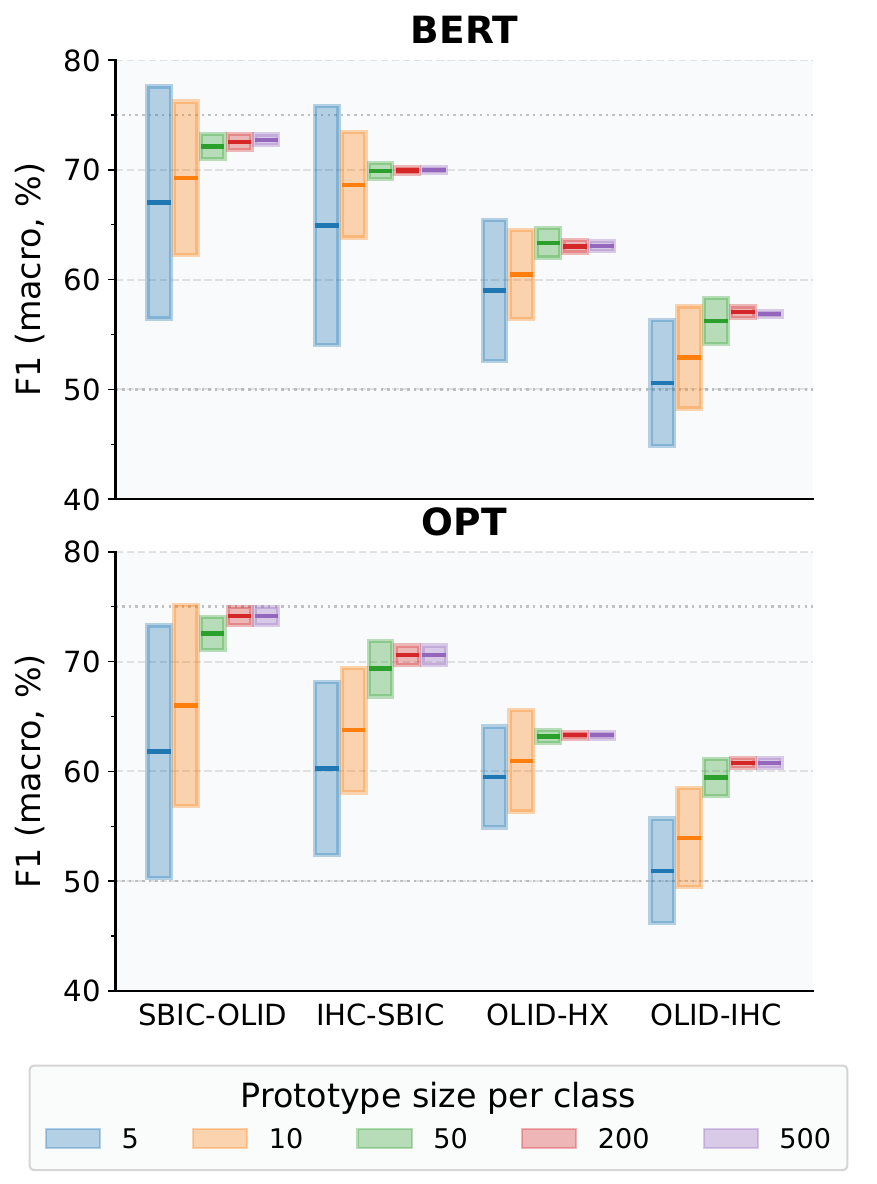}
\caption{F1-scores for BERT and OPT models across four dataset pairs (tuned source-evaluation target) with varying numbers of prototypes per class.}
  \label{fig:prototype-selection}
\end{figure}

\subsection{Prototype Selection}\label{sec:prototype-selection}
Since the performance of classification can depend on the size and selection of instances used to construct prototypes, we next experiment with several prototype configurations.  
We select four dataset pairs that do not achieve the highest score increase with prototypes in \autoref{tab:bert_opt_transfer} and evaluate performance using randomly sampled prototypes per model, repeating the process 100 times for each sample size.  
The resulting changes in F1-score across the selected pairs are illustrated in \autoref{fig:prototype-selection}.  
Overall, the F1-score obtained using classification based on only 50 prototypes is close to that achieved with 500 per-class prototype samples.  
For both OPT and BERT models, transfer performance from models tuned on implicit benchmarks improves with the number of prototypes per class.  
For the SBIC-OLID and IHC-SBIC pairs, the F1-score increases from 55\% to 70-75\% for 5-type prototype settings and stabilizes at around 50-200 examples per class.

\begin{table}[t]
\centering
\scriptsize{
\begin{tabular}{lcccc} 
\toprule
\textbf{Benchmark}  &
\textbf{Acc. Base} & \textbf{Acc. P.} & \textbf{F1 Base} & \textbf{F1 P.} \\
\midrule
\multicolumn{5}{c}{\textbf{LLaMA-1B-Guard}} \\
\midrule
HX & 66.74 & 67.95 & 64.03 & 67.59 \\
OLID & 46.86 & 71.26 & 46.36 & 60.96 \\
IHC & 52.76 & 64.28 & 52.66 & 62.27 \\
SBIC & 52.31 & 74.07 & 52.14 & 70.33 \\
\midrule
\multicolumn{5}{c}{\textbf{BLOOMZ-3B-Guard}} \\
\midrule
HX & 59.30 & 61.75 & 59.29 & 61.43 \\
OLID & 80.35 & 77.49 & 68.53 & 73.50\\
IHC & 65.64 & 63.36 & 49.49 & 60.92 \\
SBIC & 44.43 & 55.01 & 44.16 & 54.45 \\
\bottomrule
\end{tabular}}
\caption{Accuracy and macro-F1 for prototype-based classification with Guard models compared to the baseline performance of the models by benchmark.}
\label{tab:guard}
\end{table}

\subsection{Prototype Classification for Guard Models}\label{sec:guardrail-models-prototypes}

We also experiment with guard models designed for the safety evaluation of generated text.
While these models are primarily intended to detect general unsafe content rather than hate speech specifically, we aim to test whether prototypes can be effectively used to enhance their performance in classifying both implicit and explicit hate.

We use two models for these experiments: LLaMA-Guard-1B and BLOOMz-Guard-3B (see \S\ref{sec:exp-setup}).
LLaMA-Guard is released for general content safety moderation, with hate being one of the covered categories.
BLOOMz-Guard is designed to classify content into categories such as obscene, sexually explicit, identity attack, insult, and threat.
For baseline evaluation, we follow the evaluation protocols published with the respective models.
For LLaMA-Guard, we use the model’s unsafe content predictions to classify texts as hateful or not.
For BLOOMz-Guard, we classify a text as hateful if the probability for any hate-related category exceeds 0.5.

We report results in \autoref{tab:guard} for both models.
Interestingly, we find that the use of prototypes significantly enhances performance across all tested settings, with the largest macro F1-score improvements observed for SBIC with LLaMA-Guard (70.33 vs. 52.14) and for IHC with BLOOMz-Guard (60.92 vs. 49.49).
Despite having more parameters, BLOOMz-Guard achieves lower scores on SBIC (54.45), suggesting that the model is more biased toward explicit hate categories, such as those represented in OLID.

Overall, we show that the prototypes can be successfully used to enhance performance on out-of-domain data for classifiers and guardrail content moderation language models without fine-tuning.  
We also find that prototype transfer is interchangeable, indicating that out-of-domain data can be used to construct prototypes for in-domain classification, and vice versa.

\section{Early-Exiting with Prototypes}\label{sec:early-exiting-prototypes}

Next, we analyze the applicability of constructed prototypes for early exiting.  
For these experiments, we use the exiting rule defined in Eq.~\eqref{eq:main-exit}, where an exit at layer~$\ell$ is performed if the difference between the similarities of the input and the two class prototypes exceeds a threshold~$\delta$.

\subsection{Early-exiting with Prototypes}
We experiment with the fine-tuned models from \S\ref{sec:encoder-transfer-prototypes}, trained on separate datasets, four BERT and four OPT models.

\begin{table*}[t]
\footnotesize
\centering
\begin{tabular}{lcccccccc} 
\toprule
\multirow{2}{*}{\textbf{Method}}  &
\multicolumn{2}{c}{\textbf{HateXplain}} & \multicolumn{2}{c}{\textbf{OLID}} & 
\multicolumn{2}{c}{\textbf{IHC}} & \multicolumn{2}{c}{\textbf{SBIC}} \\[0.4ex]
&  \textbf{AvgExit} & \textbf{F1} & \textbf{AvgExit} & \textbf{F1} & \textbf{AvgExit} & \textbf{F1} & \textbf{AvgExit} & \textbf{F1} \\
\midrule
\multicolumn{9}{c}{\textbf{BERT}} \\
\midrule
Full model & 12 & 77.70 & 12  & 79.97 & 12 & 73.82 & 12 & 82.26 \\
Prototype-based@L12 &   12 & 77.12 & 12 & 79.91 & 12 & 73.78 & 12 & 82.37 \\
Prototype-based@L1 & 1 & 60.67 & 1 & 48.98 & 1 & 52.37 & 1 & 55.50 \\
Entropy-based & 8.50 & 76.48 & 10.81  & 79.57  & 10.44 & 75.01  & 10.10 & 82.39 \\
Patience-based & 9.15 & 77.14 & 8.62  & 51.21  & 9.03 & 48.74  & 10.18 & 87.15 \\
\midrule
Prototype-based & 9.75 & 76.64 & 10.71 & 79.89 & 10.50 & 73.87 & 10.48 & 82.14 \\
\midrule
\multicolumn{9}{c}{\textbf{OPT}} \\
\midrule
Full model & 12 & 78.62 & 12  & 80.77 & 12 & 75.75 & 12 & 83.84 \\
Prototype-based@L12 &  12 & 77.86 & 12 & 81.19 & 12 & 75.91 & 12 & 81.22 \\
Prototype-based@L1  & 1 & 53.26 & 1 & 54.55 & 1 & 56.52 & 1 & 45.87 \\
Entropy-based & 10.25 & 77.68 & 8.83  & 72.44  & 8.73 & 70.81  & 10.18 & 83.09 \\
Patience-based & 9.34 & 78.55 & 9.02  & 46.34  & 8.21 &  59.32  & 9.44 & 86.53 \\
\midrule
Prototype-based& 9.47 & 77.56 & 8.85 & 81.11 & 8.53 & 74.24 & 10.15 & 81.43 \\

\bottomrule
\end{tabular}
\caption{Early-exit performance of BERT and OPT models on four hate-speech benchmarks. We report macro-F1 scores and average exit layers. \textit{Prototype@L} denotes $\ell$-layer prototype classification shared across all samples (no per-sample exiting). \textit{Entropy-based} corresponds to DeeBERT/DeeOPT exiting, and \textit{Patience-based} to PABEE exiting. Results are shown at comparable average exit layers across baselines for fair comparison.
}
\label{tab:bert_opt_early_exiting}
\end{table*}

\paragraph{Baselines}
As the primary baseline, we use standard full-model inference, where predictions are generated after the final transformer layer. We then compare our prototype-based approach with two early-exiting methods: the entropy-based DeeBERT \citep{xin-etal-2020-deebert}, which exits once the prediction entropy falls below a predefined threshold at a given layer, and the patience-based PABEE \citep{zhou2020bert}, which exits when several consecutive layers produce consistent predictions.
In both methods, a classification gate is added to each layer and fine-tuned with OPT and BERT after model training. Since these methods were originally designed for encoder-based architectures, we adapt them for decoder-based models by similarly adding a classification gate at each layer.

\begin{figure}[b]
\centering
  \includegraphics[width=0.95\columnwidth]{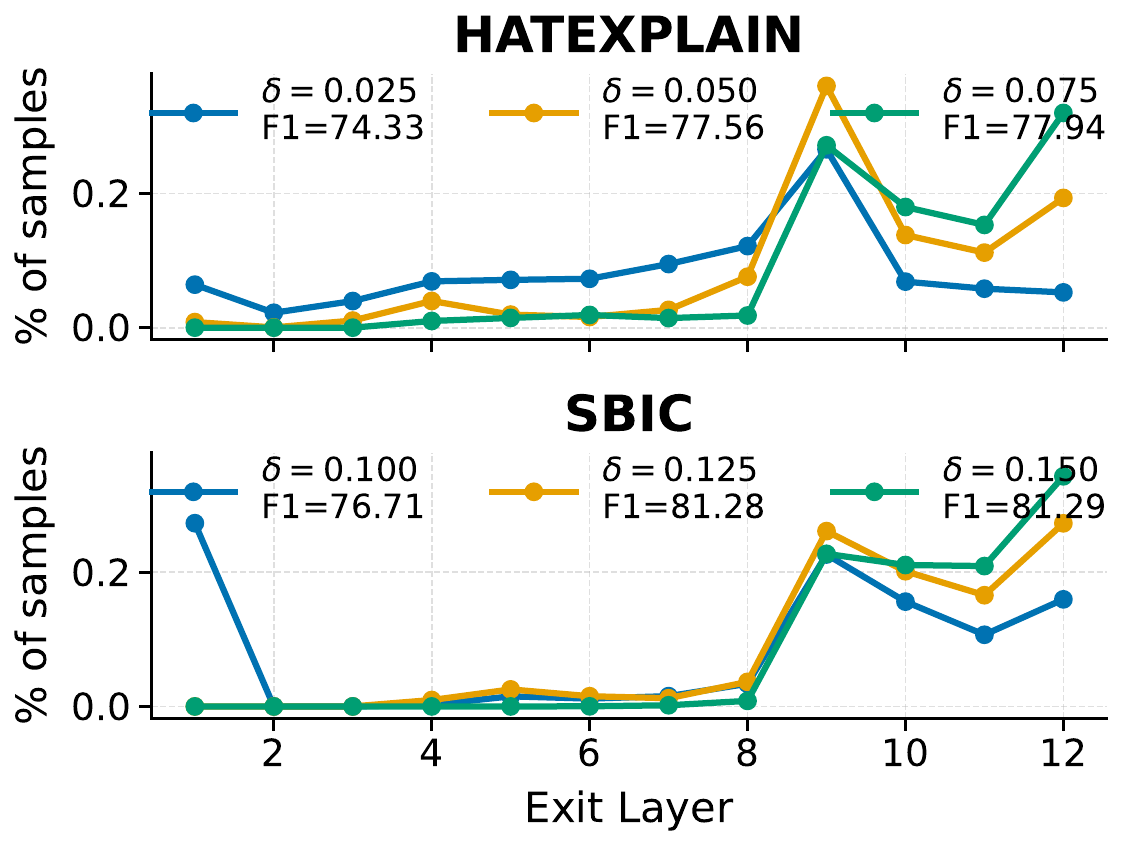}
\caption{Layer-wise proportion of samples exiting under the \textit{prototype-based} early-exit criterion for the explicit HX and implicit SBIC benchmarks using the OPT model.}
  \label{fig:layer-proportion}
\end{figure}

Although these methods rely on additional tuned parameters in their classification heads, we include them to evaluate how our parameter-free, fine-tuning-free prototype-based exiting compares to parameter-dependent exiting performance.
The total number of trained parameters increases by approximately 38k per model when using these methods, whereas the prototype-based approach requires no trained parameters and only a single threshold hyperparameter.
We use 500 examples per class to construct prototypes and employ prototypes derived from the fine-tuned objective to classify the test data from the same objective, without transfer.

\paragraph{Early Exiting for Implicit vs.\ Explicit Hate}
\autoref{tab:bert_opt_early_exiting} reports macro-F1 and average exit layers under the prototype similarity-gap criterion. 
We tune $\delta$ via a grid search over $\{0, 0.01, 0.025, 0.05, 0.075, 0.1\}$ and select the smallest value achieving approximately 20\% layer reduction while keeping macro-F1 within 1 absolute point of the full-model baseline. 
Results are averaged over 10 seeds. 
Overall, prototype-based early exiting reduces computation by about 20\% with minimal performance degradation.
On OLID, it outperforms the entropy-based DeeOPT, improving macro-F1 from 72.44\% to 81.11\%.
On HateXplain, performance remains on par with both entropy-based (DeeBERT) and patience-based (PaBEE) baselines, while patience-based methods degrade notably on OLID and IHC under similar efficiency settings.
Across architectures, trends are consistent, though implicit hate detection shows a stronger delay in exiting: BERT requires an average of 10.5 layers to match OPT’s 8.5 on SBIC.
For SBIC, where examples tend to exit later, we compare gap thresholds yielding approximately the same $\sim$20\% computational reduction as in HateXplain (see \autoref{fig:layer-proportion}). We find that most SBIC samples exit around the 9th-12th layers, whereas for HateXplain, a substantial portion still exits earlier even under comparable savings.

\begin{figure}[h]
\centering
  \includegraphics[width=0.8\columnwidth]{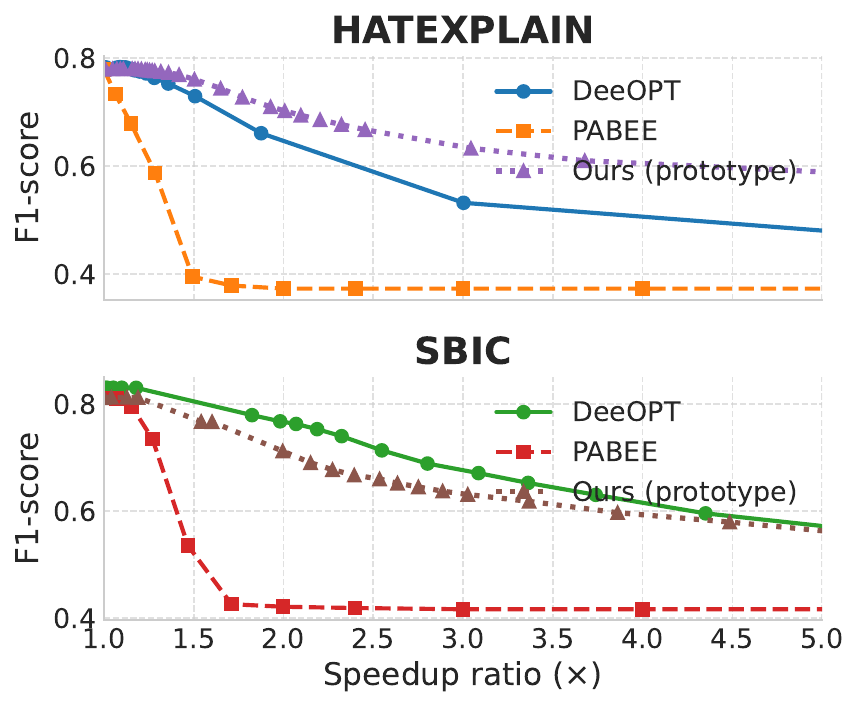}
  \caption{F1-scores vs.\ speed-up on HateXplain and SBIC for the OPT model, compared to entropy-based (DEEOPT) and patience-based (PABEE) baselines.}
  \label{fig:speed-ups}
\end{figure}

\paragraph{Inference Speed-ups with Prototypes}
We compare speedup and F1-score to complement the $\sim$20\% computational savings, where speedup is computed as $L / \bar{l}$, with $L$ denoting the total and $\bar{l}$ the average exit layer, and plot the results in \autoref{fig:speed-ups}.

Prototype-based early exiting achieves speedups comparable to entropy-based baselines while consistently outperforming the patience-based approach.
The most reliable gains, with no significant drop in F1-score, are observed for speedups below $1.5\times$ across both benchmarks.


\paragraph{Similarity gap impact}
The prototype-based exiting method requires a predefined gap parameter, $\delta$, which regulates the similarity margin. We observe that the similarity gap threshold is low (below 0.1) for exits around the 9th-10th average layers in \autoref{fig:layer-proportion}.

\begin{figure}[t]
\centering
  \includegraphics[width=0.8\columnwidth]{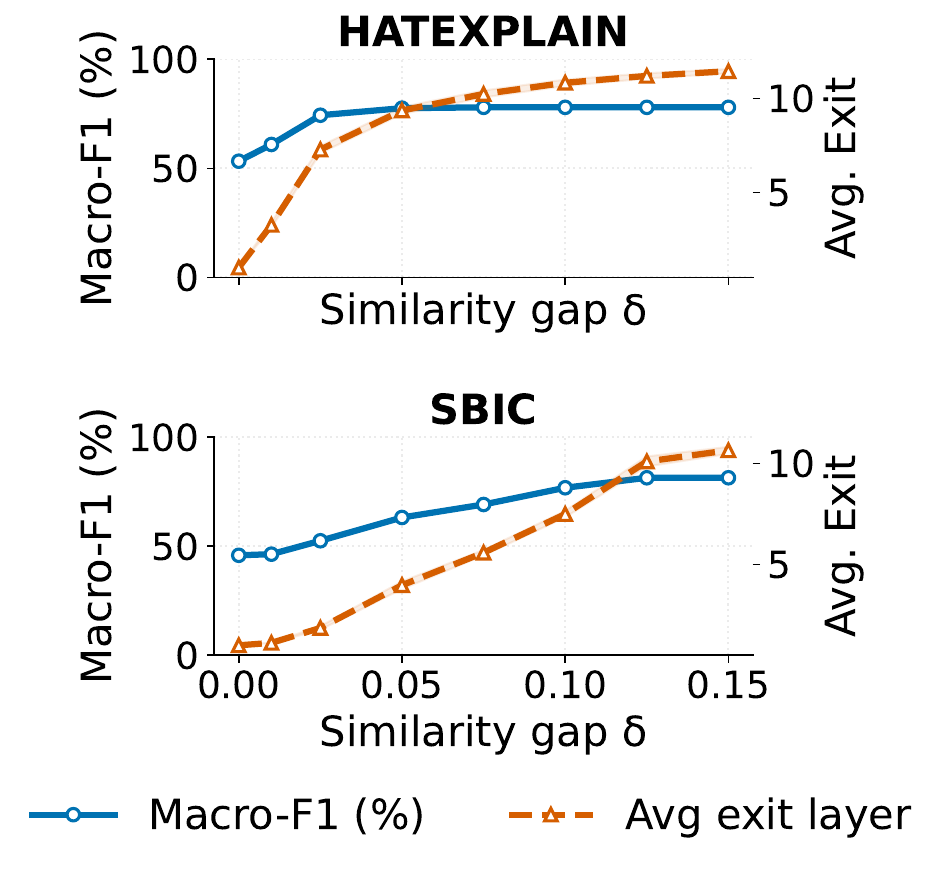}
\caption{Macro-F1 score (\%) and average exit layer across similarity gaps for the explicit HX and implicit SBIC benchmarks. Model=OPT.}
  \label{fig:sim-threshold}
\end{figure}

We further analyze the impact of the $\delta$ parameter on performance and illustrate the F1-scores per gap value for the OPT model on HateXplain and SBIC in \autoref{fig:sim-threshold}. We find that for HateXplain, the macro F1 stabilizes with a gap of 0.05 and an average exit around the 10th layer, whereas for SBIC, stabilization occurs at a gap of 0.125 with a similar average exit layer. 
These observations are consistent with the layer-wise exiting analysis and show that, for implied hate texts from SBIC that focus on implicit hate expression features, a larger gap is needed to effectively distinguish between neutral and hateful messages on the implicit SBIC benchmark. 

The relatively small similarity margin (<0.2) may stem from the substantial overlap between non-hateful and hateful texts used to construct prototypes: both categories share linguistic features, including references to group hate targets and the use of African American English expressions that occur in both hateful and non-hateful contexts.
Consequently, only minor semantic differences beyond lexical cues determine the perceived hatefulness of a message.

Overall, we find that a simple parameter-free exiting strategy based on the similarity between class prototypes can approach the performance of entropy-based baselines. 
We also observe that the exit depth required for classification is greater for implicit messages, which is consistent with the exiting behavior of baseline methods. 
For such subtle forms of hate speech, a higher gap threshold may be necessary to avoid performance degradation. The gap threshold could be further explored and calibrated on a per-layer basis.




\section{Conclusion}\label{sec:conclusion}

In this work, we present \textit{HatePrototypes}, a parameter-free approach for classifying implicit and explicit hate speech using class prototypes derived from fine-tuned language models.
We analyzed two applications: (1) prototype-based cross-domain classification and (2) prototype-guided early exiting.

Our results show that prototype representations substantially improve out-of-domain performance without degrading in-domain accuracy, with prototypes constructed from as few as 50 examples per class.
We also find that prototype-based classification enhances the performance of open safety moderation models on both implicit and explicit benchmarks.
Next, we find that prototype similarity can be effectively used to support anytime prediction at different layers in LMs, with differences varying depending on the level of explicitness of the text.

To facilitate further research, we will release the HatePrototypes framework and prototype resources to support cross-model and cross-dataset analysis. 
This contribution will enable a systematic examination of how hate-related representations differ across model architectures, layers, and benchmarks, helping to identify the limitations of current systems in real-world moderation. 
By analyzing examples with low prototype similarity or inconsistent predictions, researchers can detect ambiguous or underrepresented cases of hate and use these findings to guide the development of more comprehensive and balanced hate speech datasets.

\section*{Limitations}
While efficient, \textit{HatePrototypes}, like other early-exiting techniques and methods for model acceleration, can lead to lower scores on out-of-domain test data compared to full-model baselines. 
However, the exiting threshold can be further calibrated at the layer-wise level to significantly reduce the performance gap introduced by early exiting. 

Similarly, \textit{HatePrototypes} transfer relies on the hidden states of fine-tuned LMs, which may be insufficient to achieve the same performance as baselines whose parameters are directly optimized on the target benchmark.
In this work, we focus on hidden representations of texts derived from LMs. Future research could extend the proposed prototype-based framework to multiview representations within multimodal architectures, where prototypes are jointly constructed from text and additional modalities.

Another limitation lies in the need for reliable annotations in implicit-hate benchmarks. We experiment with widely used corpora for hate-speech classification; however, the subtle nature of implicit hate often results in low annotation reproducibility, as judgments can vary considerably across annotators and cultural contexts. Overall, the development of new resources remains complex and may require fine-grained annotation schemes. 
The released prototype-based framework could be used to pre-annotate implicit-hate data and highlight the most ambiguous cases for human review.

Finally, prototype-based early exiting could also be leveraged for interpretability to explore how model depth reflects the subtlety of a classified instance and to understand the processing depth required for a correct prediction.

\section*{Ethical Considerations}
This study investigates the use of prototype-based methods for hate speech detection in cross-task transfer and early-exiting settings. 
We rely on four publicly available benchmarks, including fine-grained annotations from the IHC dataset, and use them in accordance with their respective licenses.

We evaluate performance using classification and efficiency metrics; however, we do not examine the potential effects of early exiting on biases in texts targeting different minorities or protected groups. 
Such aspects remain outside the scope of this study. 

Future research could extend this analysis to identity- and group-directed hate targets, with particular attention to ensuring comparable performance across groups or by constructing group-specific prototypes, for instance, using hateful texts targeting a given group and neutral texts mentioning the same group.

We release the code to support the following intended uses: (1) studying transferability in hate speech models, and (2) enabling early exiting for model acceleration in hate speech detection and related research.\footnote{\url{https://github.com/upunaprosk/hate-prototypes}}

\paragraph{Acknowledgments}
This work was performed using HPC resources from GENCI-IDRIS (Grant 2025-AD011014384).
Additionally, this research was supported by the ANR project Diké (No. ANR-21-CE23-0026-02).

\section{Bibliographical References}\label{sec:reference}

\bibliographystyle{lrec2026-natbib}
\bibliography{lrec2026-example}

\begin{thebibliography}{4}
\expandafter\ifx\csname natexlab\endcsname\relax\def\natexlab#1{#1}\fi

\bibitem[{ElSherief et~al.(2021)ElSherief, Ziems, Muchlinski, Anupindi, Seybolt, De~Choudhury, and Yang}]{elsherief2021latent}
ElSherief, Mai and Ziems, Caleb and Muchlinski, David and Anupindi, Vaishnavi and Seybolt, Jordyn and De Choudhury, Munmun and Yang, Diyi. 2021.
\newblock \href {https://doi.org/10.18653/v1/2021.emnlp-main.29} {\emph{Latent Hatred: A Benchmark for Understanding Implicit Hate Speech}}.
\newblock Association for Computational Linguistics.

\bibitem[{Mathew et~al.(2021)Mathew, Saha, Yimam, Biemann, Goyal, and Mukherjee}]{mathew2021hatexplain}
Mathew, Binny and Saha, Punyajoy and Yimam, Seid Muhie and Biemann, Chris and Goyal, Pawan and Mukherjee, Animesh. 2021.
\newblock \href {https://doi.org/10.1609/aaai.v35i17.17745} {\emph{HateXplain: A Benchmark Dataset for Explainable Hate Speech Detection}}.

\bibitem[{Sap et~al.(2020)Sap, Gabriel, Qin, Jurafsky, Smith, and Choi}]{sap-etal-2020-social}
Sap, Maarten and Gabriel, Saadia and Qin, Lianhui and Jurafsky, Dan and Smith, Noah A. and Choi, Yejin. 2020.
\newblock \href {https://doi.org/10.18653/v1/2020.acl-main.486} {\emph{Social Bias Frames: Reasoning about Social and Power Implications of Language}}.
\newblock Association for Computational Linguistics.

\bibitem[{Zampieri et~al.(2019)Zampieri, Malmasi, Nakov, Rosenthal, Farra, and Kumar}]{zampieri2019predicting}
Zampieri, Marcos and Malmasi, Shervin and Nakov, Preslav and Rosenthal, Sara and Farra, Noura and Kumar, Ritesh. 2019.
\newblock \href {https://doi.org/10.18653/v1/N19-1144} {\emph{Predicting the Type and Target of Offensive Posts in Social Media}}.
\newblock Association for Computational Linguistics.

\end{thebibliography}


\begin{thebibliography}{44}
\expandafter\ifx\csname natexlab\endcsname\relax\def\natexlab#1{#1}\fi

\bibitem[{Ahn et~al.(2024)Ahn, Kim, Kim, and Han}]{ahn-etal-2024-sharedcon}
Hyeseon Ahn, Youngwook Kim, Jungin Kim, and Yo-Sub Han. 2024.
\newblock \href {https://doi.org/10.18653/v1/2024.findings-acl.622} {{S}hared{C}on: Implicit hate speech detection using shared semantics}.
\newblock In \emph{Findings of the Association for Computational Linguistics: ACL 2024}, pages 10444--10455, Bangkok, Thailand. Association for Computational Linguistics.

\bibitem[{Albanyan et~al.(2023)Albanyan, Hassan, and Blanco}]{albanyan2023finding}
Abdullah Albanyan, Ahmed Hassan, and Eduardo Blanco. 2023.
\newblock \href {https://doi.org/10.18653/v1/2023.emnlp-main.855} {Finding authentic counterhate arguments: A case study with public figures}.
\newblock In \emph{Proceedings of the 2023 Conference on Empirical Methods in Natural Language Processing}, pages 13862--13876, Singapore. Association for Computational Linguistics.

\bibitem[{Arango et~al.(2019)Arango, P{\'e}rez, and Poblete}]{arango2019hate}
A.~Arango, J.~P{\'e}rez, and B.~Poblete. 2019.
\newblock \href {https://doi.org/10.1145/3331184.3331262} {Hate speech detection is not as easy as you may think: A closer look at model evaluation}.
\newblock In \emph{Proceedings of the 42nd International ACM SIGIR Conference on Research and Development in Information Retrieval}, pages 45--54. ACM.

\bibitem[{Bae et~al.(2023)Bae, Ko, Song, and Yun}]{bae-etal-2023-fast}
Sangmin Bae, Jongwoo Ko, Hwanjun Song, and Se-Young Yun. 2023.
\newblock \href {https://doi.org/10.18653/v1/2023.emnlp-main.362} {Fast and robust early-exiting framework for autoregressive language models with synchronized parallel decoding}.
\newblock In \emph{Proceedings of the 2023 Conference on Empirical Methods in Natural Language Processing}, pages 5910--5924, Singapore. Association for Computational Linguistics.

\bibitem[{Dale et~al.(2021)Dale, Voronov, Dementieva, Logacheva, Kozlova, Semenov, and Panchenko}]{dale-etal-2021-text}
David Dale, Anton Voronov, Daryna Dementieva, Varvara Logacheva, Olga Kozlova, Nikita Semenov, and Alexander Panchenko. 2021.
\newblock \href {https://doi.org/10.18653/v1/2021.emnlp-main.629} {Text detoxification using large pre-trained neural models}.
\newblock In \emph{Proceedings of the 2021 Conference on Empirical Methods in Natural Language Processing}, pages 7979--7996, Online and Punta Cana, Dominican Republic. Association for Computational Linguistics.

\bibitem[{Devlin et~al.(2019)Devlin, Chang, Lee, and Toutanova}]{devlin2019bert}
Jacob Devlin, Ming-Wei Chang, Kenton Lee, and Kristina Toutanova. 2019.
\newblock \href {https://doi.org/10.18653/v1/N19-1423} {{BERT}: Pre-training of deep bidirectional transformers for language understanding}.
\newblock In \emph{Proceedings of the 2019 Conference of the North {A}merican Chapter of the Association for Computational Linguistics: Human Language Technologies, Volume 1 (Long and Short Papers)}, pages 4171--4186, Minneapolis, Minnesota. Association for Computational Linguistics.

\bibitem[{D{\'\i}az and Hecht-Felella(2021)}]{diaz2021double}
{\'A}ngel D{\'\i}az and Laura Hecht-Felella. 2021.
\newblock \href {https://www.brennancenter.org/our-work/research-reports/double-standards-social-media-content-moderation} {Double standards in social media content moderation}.
\newblock \emph{Brennan Center for Justice at New York University School of Law}, pages 1--23.

\bibitem[{Elhoushi et~al.(2024)Elhoushi, Shrivastava, Liskovich, Hosmer, Wasti, Lai, Mahmoud, Acun, Agarwal, Roman, Aly, Chen, and Wu}]{elhoushi2024layerskip}
Mostafa Elhoushi, Akshat Shrivastava, Diana Liskovich, Basil Hosmer, Bram Wasti, Liangzhen Lai, Anas Mahmoud, Bilge Acun, Saurabh Agarwal, Ahmed Roman, Ahmed Aly, Beidi Chen, and Carole-Jean Wu. 2024.
\newblock \href {https://doi.org/10.18653/v1/2024.acl-long.681} {{L}ayer{S}kip: Enabling early exit inference and self-speculative decoding}.
\newblock In \emph{Proceedings of the 62nd Annual Meeting of the Association for Computational Linguistics (Volume 1: Long Papers)}, pages 12622--12642, Bangkok, Thailand. Association for Computational Linguistics.

\bibitem[{Görmez et~al.(2022)Görmez, Dasari, and Koyuncu}]{gormez20222}
Alperen Görmez, Venkat~R. Dasari, and Erdem Koyuncu. 2022.
\newblock \href {https://doi.org/10.1109/IJCNN55064.2022.9891952} {E2cm: Early exit via class means for efficient supervised and unsupervised learning}.
\newblock In \emph{2022 International Joint Conference on Neural Networks (IJCNN)}, pages 1--8.

\bibitem[{Hartvigsen et~al.(2022)Hartvigsen, Gabriel, Palangi, Sap, Ray, and Kamar}]{hartvigsen-etal-2022-toxigen}
Thomas Hartvigsen, Saadia Gabriel, Hamid Palangi, Maarten Sap, Dipankar Ray, and Ece Kamar. 2022.
\newblock \href {https://doi.org/10.18653/v1/2022.acl-long.234} {{T}oxi{G}en: A large-scale machine-generated dataset for adversarial and implicit hate speech detection}.
\newblock In \emph{Proceedings of the 60th Annual Meeting of the Association for Computational Linguistics (Volume 1: Long Papers)}, pages 3309--3326, Dublin, Ireland. Association for Computational Linguistics.

\bibitem[{He et~al.(2024)He, Zhang, Ding, Miao, Zhao, Hu, and Cao}]{he20243}
Jianing He, Qi~Zhang, Weiping Ding, Duoqian Miao, Jun Zhao, Liang Hu, and Longbing Cao. 2024.
\newblock \href {http://arxiv.org/abs/2402.05948} {De$^3$-bert: Distance-enhanced early exiting for bert based on prototypical networks}.

\bibitem[{Inan et~al.(2023)Inan, Upasani, Chi, Rungta, Iyer, Mao, Tontchev, Hu, Fuller, Testuggine, and Khabsa}]{inan2023llama}
Hakan Inan, Kartikeya Upasani, Jianfeng Chi, Rashi Rungta, Krithika Iyer, Yuning Mao, Michael Tontchev, Qing Hu, Brian Fuller, Davide Testuggine, and Madian Khabsa. 2023.
\newblock \href {http://arxiv.org/abs/2312.06674} {Llama guard: Llm-based input-output safeguard for human-ai conversations}.

\bibitem[{Jafari et~al.(2023)Jafari, Li, Rajapaksha, Farahbakhsh, and Crespi}]{jafari2023fine}
Amir~Reza Jafari, Guanlin Li, Praboda Rajapaksha, Reza Farahbakhsh, and Noel Crespi. 2023.
\newblock \href {https://doi.org/10.1109/ACCESS.2023.3318863} {Fine-grained emotions influence on implicit hate speech detection}.
\newblock \emph{IEEE Access}, 11:105330--105343.

\bibitem[{Jin et~al.(2024)Jin, Wanner, and Shvets}]{jin-etal-2024-gpt}
Yiping Jin, Leo Wanner, and Alexander Shvets. 2024.
\newblock \href {https://aclanthology.org/2024.lrec-main.694/} {{GPT}-{H}ate{C}heck: Can {LLM}s write better functional tests for hate speech detection?}
\newblock In \emph{Proceedings of the 2024 Joint International Conference on Computational Linguistics, Language Resources and Evaluation (LREC-COLING 2024)}, pages 7867--7885, Torino, Italia. ELRA and ICCL.

\bibitem[{Khurana et~al.(2022)Khurana, Vermeulen, Nalisnick, Van~Noorloos, and Fokkens}]{khurana-etal-2022-hate}
Urja Khurana, Ivar Vermeulen, Eric Nalisnick, Marloes Van~Noorloos, and Antske Fokkens. 2022.
\newblock \href {https://doi.org/10.18653/v1/2022.woah-1.17} {Hate speech criteria: A modular approach to task-specific hate speech definitions}.
\newblock In \emph{Proceedings of the Sixth Workshop on Online Abuse and Harms (WOAH)}, pages 176--191, Seattle, Washington (Hybrid). Association for Computational Linguistics.

\bibitem[{Kim et~al.(2022)Kim, Park, and Han}]{kim-etal-2022-generalizable}
Youngwook Kim, Shinwoo Park, and Yo-Sub Han. 2022.
\newblock \href {https://aclanthology.org/2022.coling-1.579/} {Generalizable implicit hate speech detection using contrastive learning}.
\newblock In \emph{Proceedings of the 29th International Conference on Computational Linguistics}, pages 6667--6679, Gyeongju, Republic of Korea. International Committee on Computational Linguistics.

\bibitem[{Lees et~al.(2022)Lees, Tran, Tay, Sorensen, Gupta, Metzler, and Vasserman}]{lees2022new}
Alyssa Lees, Vinh~Q. Tran, Yi~Tay, Jeffrey Sorensen, Jai Gupta, Donald Metzler, and Lucy Vasserman. 2022.
\newblock \href {https://doi.org/10.1145/3534678.3539147} {A new generation of perspective api: Efficient multilingual character-level transformers}.
\newblock In \emph{Proceedings of the 28th ACM SIGKDD Conference on Knowledge Discovery and Data Mining}, KDD '22, page 3197–3207, New York, NY, USA. Association for Computing Machinery.

\bibitem[{Leite et~al.(2023)Leite, Scarton, and Silva}]{leite-etal-2023-noisy}
Jo{\~a}o Leite, Carolina Scarton, and Diego Silva. 2023.
\newblock \href {https://aclanthology.org/2023.ranlp-1.68/} {Noisy self-training with data augmentations for offensive and hate speech detection tasks}.
\newblock In \emph{Proceedings of the 14th International Conference on Recent Advances in Natural Language Processing}, pages 631--640, Varna, Bulgaria. INCOMA Ltd., Shoumen, Bulgaria.

\bibitem[{Liu et~al.(2020)Liu, Zhou, Wang, Zhao, Deng, and Ju}]{liu2020fastbert}
Weijie Liu, Peng Zhou, Zhiruo Wang, Zhe Zhao, Haotang Deng, and Qi~Ju. 2020.
\newblock \href {https://doi.org/10.18653/v1/2020.acl-main.537} {{F}ast{BERT}: a self-distilling {BERT} with adaptive inference time}.
\newblock In \emph{Proceedings of the 58th Annual Meeting of the Association for Computational Linguistics}, pages 6035--6044, Online. Association for Computational Linguistics.

\bibitem[{Liu et~al.(2022)Liu, Sun, He, Wu, Wu, Zhang, Jiang, Cao, Huang, and Qiu}]{liu2021towards}
Xiangyang Liu, Tianxiang Sun, Junliang He, Jiawen Wu, Lingling Wu, Xinyu Zhang, Hao Jiang, Zhao Cao, Xuanjing Huang, and Xipeng Qiu. 2022.
\newblock \href {https://doi.org/10.18653/v1/2022.naacl-main.240} {Towards efficient {NLP}: A standard evaluation and a strong baseline}.
\newblock In \emph{Proceedings of the 2022 Conference of the North American Chapter of the Association for Computational Linguistics: Human Language Technologies}, pages 3288--3303, Seattle, United States. Association for Computational Linguistics.

\bibitem[{MacAvaney et~al.(2019)MacAvaney, Yao, Yang, Russell, Goharian, and Frieder}]{macavaney2019hate}
Sean MacAvaney, Hao-Ren Yao, Eugene Yang, Katina Russell, Nazli Goharian, and Ophir Frieder. 2019.
\newblock \href {https://doi.org/10.1371/journal.pone.0221152} {Hate speech detection: Challenges and solutions}.
\newblock \emph{PLOS ONE}, 14(8):1--16.

\bibitem[{Müller and Schwarz(2020)}]{muller2021fanning}
Karsten Müller and Carlo Schwarz. 2020.
\newblock \href {https://doi.org/10.1093/jeea/jvaa045} {Fanning the flames of hate: Social media and hate crime}.
\newblock \emph{Journal of the European Economic Association}, 19(4):2131--2167.

\bibitem[{Nadeem et~al.(2021)Nadeem, Bethke, and Reddy}]{nadeem-etal-2021-stereoset}
Moin Nadeem, Anna Bethke, and Siva Reddy. 2021.
\newblock \href {https://doi.org/10.18653/v1/2021.acl-long.416} {{S}tereo{S}et: Measuring stereotypical bias in pretrained language models}.
\newblock In \emph{Proceedings of the 59th Annual Meeting of the Association for Computational Linguistics and the 11th International Joint Conference on Natural Language Processing (Volume 1: Long Papers)}, pages 5356--5371, Online. Association for Computational Linguistics.

\bibitem[{Nangia et~al.(2020)Nangia, Vania, Bhalerao, and Bowman}]{nangia-etal-2020-crows}
Nikita Nangia, Clara Vania, Rasika Bhalerao, and Samuel~R. Bowman. 2020.
\newblock \href {https://doi.org/10.18653/v1/2020.emnlp-main.154} {{C}row{S}-pairs: A challenge dataset for measuring social biases in masked language models}.
\newblock In \emph{Proceedings of the 2020 Conference on Empirical Methods in Natural Language Processing (EMNLP)}, pages 1953--1967, Online. Association for Computational Linguistics.

\bibitem[{Ocampo et~al.(2023)Ocampo, Sviridova, Cabrio, and Villata}]{ocampo2020detecting}
Nicol{\'a}s~Benjam{\'i}n Ocampo, Ekaterina Sviridova, Elena Cabrio, and Serena Villata. 2023.
\newblock \href {https://doi.org/10.18653/v1/2023.eacl-main.147} {An in-depth analysis of implicit and subtle hate speech messages}.
\newblock In \emph{Proceedings of the 17th Conference of the European Chapter of the Association for Computational Linguistics}, pages 1997--2013, Dubrovnik, Croatia. Association for Computational Linguistics.

\bibitem[{Oikawa et~al.(2022)Oikawa, Nakayama, and Murakami}]{oikawa2022stacking}
Yuto Oikawa, Yuki Nakayama, and Koji Murakami. 2022.
\newblock \href {https://doi.org/10.18653/v1/2022.emnlp-industry.58} {A stacking-based efficient method for toxic language detection on live streaming chat}.
\newblock In \emph{Proceedings of the 2022 Conference on Empirical Methods in Natural Language Processing: Industry Track}, pages 571--578, Abu Dhabi, UAE. Association for Computational Linguistics.

\bibitem[{Pachinger et~al.(2023)Pachinger, Hanbury, Neidhardt, and Planitzer}]{pachinger2023toward}
Pia Pachinger, Allan Hanbury, Julia Neidhardt, and Anna Planitzer. 2023.
\newblock \href {https://doi.org/10.18653/v1/2023.c3nlp-1.11} {Toward disambiguating the definitions of abusive, offensive, toxic, and uncivil comments}.
\newblock In \emph{Proceedings of the First Workshop on Cross-Cultural Considerations in NLP (C3NLP)}, pages 107--113, Dubrovnik, Croatia. Association for Computational Linguistics.

\bibitem[{Rahmath~P et~al.(2024)Rahmath~P, Srivastava, Chaurasia, Pacheco, and Couto}]{rahmath2024early}
Haseena Rahmath~P, Vishal Srivastava, Kuldeep Chaurasia, Roberto~G. Pacheco, and Rodrigo~S. Couto. 2024.
\newblock \href {https://doi.org/10.1145/3698767} {Early-exit deep neural network - a comprehensive survey}.
\newblock \emph{ACM Comput. Surv.}, 57(3).

\bibitem[{R{\"o}ttger et~al.(2021)R{\"o}ttger, Vidgen, Nguyen, Waseem, Margetts, and Pierrehumbert}]{rottger-etal-2021-hatecheck}
Paul R{\"o}ttger, Bertie Vidgen, Dong Nguyen, Zeerak Waseem, Helen Margetts, and Janet Pierrehumbert. 2021.
\newblock \href {https://doi.org/10.18653/v1/2021.acl-long.4} {Hatecheck: Functional tests for hate speech detection models}.
\newblock In \emph{Proceedings of the 59th Annual Meeting of the Association for Computational Linguistics and the 11th International Joint Conference on Natural Language Processing (Volume 1: Long Papers)}, pages 41--58, Online. Association for Computational Linguistics.

\bibitem[{Schmidt and Wiegand(2017)}]{schmidt2017survey}
Anna Schmidt and Michael Wiegand. 2017.
\newblock \href {https://doi.org/10.18653/v1/W17-1101} {A survey on hate speech detection using natural language processing}.
\newblock In \emph{Proceedings of the Fifth International Workshop on Natural Language Processing for Social Media}, pages 1--10, Valencia, Spain. Association for Computational Linguistics.

\bibitem[{Snell et~al.(2017)Snell, Swersky, and Zemel}]{snell2017prototypical}
Jake Snell, Kevin Swersky, and Richard Zemel. 2017.
\newblock \href {https://proceedings.neurips.cc/paper_files/paper/2017/file/cb8da6767461f2812ae4290eac7cbc42-Paper.pdf} {Prototypical networks for few-shot learning}.
\newblock In \emph{Advances in Neural Information Processing Systems}, volume~30. Curran Associates, Inc.

\bibitem[{Sridhar and Yang(2022)}]{sridhar2022explaining}
Rohit Sridhar and Diyi Yang. 2022.
\newblock \href {https://doi.org/10.18653/v1/2022.naacl-main.59} {Explaining toxic text via knowledge enhanced text generation}.
\newblock In \emph{Proceedings of the 2022 Conference of the North American Chapter of the Association for Computational Linguistics: Human Language Technologies}, pages 811--826, Seattle, United States. Association for Computational Linguistics.

\bibitem[{Tang et~al.(2024)Tang, Zhu, Li, Appalaraju, Mahadevan, and Manmatha}]{tang-etal-2024-deed}
Peng Tang, Pengkai Zhu, Tian Li, Srikar Appalaraju, Vijay Mahadevan, and R.~Manmatha. 2024.
\newblock \href {https://doi.org/10.18653/v1/2024.findings-naacl.9} {{DEED}: Dynamic early exit on decoder for accelerating encoder-decoder transformer models}.
\newblock In \emph{Findings of the Association for Computational Linguistics: NAACL 2024}, pages 116--131, Mexico City, Mexico. Association for Computational Linguistics.

\bibitem[{Tonneau et~al.(2025)Tonneau, Liu, Malhotra, Hale, Fraiberger, Orozco-Olvera, and R{\"o}ttger}]{tonneau-etal-2025-hateday}
Manuel Tonneau, Diyi Liu, Niyati Malhotra, Scott~A. Hale, Samuel Fraiberger, Victor Orozco-Olvera, and Paul R{\"o}ttger. 2025.
\newblock \href {https://doi.org/10.18653/v1/2025.acl-long.115} {{H}ate{D}ay: Insights from a global hate speech dataset representative of a day on {T}witter}.
\newblock In \emph{Proceedings of the 63rd Annual Meeting of the Association for Computational Linguistics (Volume 1: Long Papers)}, pages 2297--2321, Vienna, Austria. Association for Computational Linguistics.

\bibitem[{Treviso et~al.(2023)Treviso, Lee, Ji, van Aken, Cao, Ciosici, Hassid, Heafield, Hooker, Raffel, Martins, Martins, Forde, Milder, Simpson, Slonim, Dodge, Strubell, Balasubramanian, Derczynski, Gurevych, and Schwartz}]{treviso2023efficient}
Marcos Treviso, Ji-Ung Lee, Tianchu Ji, Betty van Aken, Qingqing Cao, Manuel~R. Ciosici, Michael Hassid, Kenneth Heafield, Sara Hooker, Colin Raffel, Pedro~H. Martins, Andr{\'e} F.~T. Martins, Jessica~Zosa Forde, Peter Milder, Edwin Simpson, Noam Slonim, Jesse Dodge, Emma Strubell, Niranjan Balasubramanian, Leon Derczynski, Iryna Gurevych, and Roy Schwartz. 2023.
\newblock \href {https://doi.org/10.1162/tacl_a_00577} {Efficient methods for natural language processing: A survey}.
\newblock \emph{Transactions of the Association for Computational Linguistics}, 11:826--860.

\bibitem[{Vidgen and Derczynski(2020)}]{vidgen-derczynski-2020}
Bertie Vidgen and Leon Derczynski. 2020.
\newblock \href {https://doi.org/10.1371/journal.pone.0243300} {Directions in abusive language training data, a systematic review: Garbage in, garbage out}.
\newblock \emph{PLOS ONE}, 15(12):e0243300.

\bibitem[{Waseem and Hovy(2016)}]{waseem-hovy-2016-hateful}
Zeerak Waseem and Dirk Hovy. 2016.
\newblock \href {https://doi.org/10.18653/v1/N16-2013} {Hateful symbols or hateful people? predictive features for hate speech detection on {T}witter}.
\newblock In \emph{Proceedings of the {NAACL} Student Research Workshop}, pages 88--93, San Diego, California. Association for Computational Linguistics.

\bibitem[{Wiegand et~al.(2019)Wiegand, Ruppenhofer, and Kleinbauer}]{wiegand2019detection}
Michael Wiegand, Josef Ruppenhofer, and Thomas Kleinbauer. 2019.
\newblock \href {https://doi.org/10.18653/v1/N19-1060} {{D}etection of {A}busive {L}anguage: the {P}roblem of {B}iased {D}atasets}.
\newblock In \emph{Proceedings of the 2019 Conference of the North {A}merican Chapter of the Association for Computational Linguistics: Human Language Technologies, Volume 1 (Long and Short Papers)}, pages 602--608, Minneapolis, Minnesota. Association for Computational Linguistics.

\bibitem[{Xie et~al.(2023)Xie, Vosoughi, and Hassanpour}]{xie-etal-2023-proto}
Sean Xie, Soroush Vosoughi, and Saeed Hassanpour. 2023.
\newblock \href {https://doi.org/10.18653/v1/2023.findings-emnlp.261} {Proto-lm: A prototypical network-based framework for built-in interpretability in large language models}.
\newblock In \emph{Findings of the Association for Computational Linguistics: EMNLP 2023}, pages 3964--3979, Singapore. Association for Computational Linguistics.

\bibitem[{Xin et~al.(2020)Xin, Tang, Lee, Yu, and Lin}]{xin-etal-2020-deebert}
Ji~Xin, Raphael Tang, Jaejun Lee, Yaoliang Yu, and Jimmy Lin. 2020.
\newblock \href {https://doi.org/10.18653/v1/2020.acl-main.204} {{D}ee{BERT}: Dynamic early exiting for accelerating {BERT} inference}.
\newblock In \emph{Proceedings of the 58th Annual Meeting of the Association for Computational Linguistics}, pages 2246--2251, Online. Association for Computational Linguistics.

\bibitem[{Yang et~al.(2023)Yang, Grenon-Godbout, and Rabbany}]{yang2023toxbuster}
Zachary Yang, Nicolas Grenon-Godbout, and Reihaneh Rabbany. 2023.
\newblock \href {https://doi.org/10.18653/v1/2023.findings-emnlp.663} {Towards detecting contextual real-time toxicity for in-game chat}.
\newblock In \emph{Findings of the Association for Computational Linguistics: EMNLP 2023}, pages 9894--9906, Singapore. Association for Computational Linguistics.

\bibitem[{Zhang et~al.(2022)Zhang, Roller, Goyal, Artetxe, Chen, Chen, Dewan, Diab, Li, Lin, Mihaylov, Ott, Shleifer, Shuster, Simig, Koura, Sridhar, Wang, and Zettlemoyer}]{zhang2022opt}
Susan Zhang, Stephen Roller, Naman Goyal, Mikel Artetxe, Moya Chen, Shuohui Chen, Christopher Dewan, Mona Diab, Xian Li, Xi~Victoria Lin, Todor Mihaylov, Myle Ott, Sam Shleifer, Kurt Shuster, Daniel Simig, Punit~Singh Koura, Anjali Sridhar, Tianlu Wang, and Luke Zettlemoyer. 2022.
\newblock \href {http://arxiv.org/abs/2205.01068} {Opt: Open pre-trained transformer language models}.

\bibitem[{Zhao et~al.(2025)Zhao, Wang, Wang, Zhao, He, and Hou}]{zhao-etal-2025-explicit}
Yachao Zhao, Bo~Wang, Yan Wang, Dongming Zhao, Ruifang He, and Yuexian Hou. 2025.
\newblock \href {https://doi.org/10.18653/v1/2025.findings-acl.1} {Explicit vs. implicit: Investigating social bias in large language models through self-reflection}.
\newblock In \emph{Findings of the Association for Computational Linguistics: ACL 2025}, pages 1--12, Vienna, Austria. Association for Computational Linguistics.

\bibitem[{Zhou et~al.(2020)Zhou, Xu, Ge, McAuley, Xu, and Wei}]{zhou2020bert}
Wangchunshu Zhou, Canwen Xu, Tao Ge, Julian McAuley, Ke~Xu, and Furu Wei. 2020.
\newblock \href {https://proceedings.neurips.cc/paper_files/paper/2020/file/d4dd111a4fd973394238aca5c05bebe3-Paper.pdf} {Bert loses patience: Fast and robust inference with early exit}.
\newblock In \emph{Advances in Neural Information Processing Systems}, volume~33, pages 18330--18341. Curran Associates, Inc.

\end{thebibliography}

\section{Language Resource References}
\label{lr:ref}
\bibliographystylelanguageresource{lrec2026-natbib}
\bibliographylanguageresource{languageresource}

\end{document}